\documentclass[journal,draftclsnofoot]{IEEEtran}

\usepackage{cite}
\usepackage{amsmath,amssymb,amsfonts}
\usepackage{algorithmic}
\usepackage{graphicx}
\usepackage{textcomp}
\usepackage{xcolor}
\usepackage{lipsum}
\usepackage{float}
\usepackage{booktabs}
\usepackage[per-mode=symbol]{siunitx}
\usepackage{subcaption}
\usepackage{multirow}
\usepackage{hyperref}
\usepackage{threeparttable}
\usepackage{balance}

\usepackage{textcase}
\usepackage[tablename=TABLE]{caption}
\DeclareCaptionTextFormat{up}{\MakeTextUppercase{#1}}
\usepackage{tikz}
\usepackage{textcomp}
\usepackage{hyperref}
\usepackage{lipsum}

\newcommand\copyrighttext{%
  \footnotesize \textcopyright 2025 IEEE. Personal use of this material is permitted.  Permission from IEEE must be obtained for all other uses, in any current or future media, including reprinting/republishing this material for advertising or promotional purposes, creating new collective works, for resale or redistribution to servers or lists, or reuse of any copyrighted component of this work in other works.}
\newcommand\copyrightnotice{%
\begin{tikzpicture}[remember picture,overlay]
\node[anchor=south,yshift=10pt] at (current page.south) {\fbox{\parbox{\dimexpr\textwidth-\fboxsep-\fboxrule\relax}{\copyrighttext}}};
\end{tikzpicture}%
}

\def\BibTeX{{\rm B\kern-.05em{\sc i\kern-.025em b}\kern-.08em
    T\kern-.1667em\lower.7ex\hbox{E}\kern-.125emX}}

\begin{document}
%
\title{On-device Anomaly Detection in Conveyor Belt Operations}

\author{\IEEEauthorblockN{Luciano Sebastian Martinez-Rau\IEEEauthorrefmark{1}\IEEEauthorrefmark{2},~\IEEEmembership{Member,~IEEE},
Yuxuan Zhang\IEEEauthorrefmark{1},~\IEEEmembership{Graduate Student Member,~IEEE}, \\
Bengt Oelmann\IEEEauthorrefmark{1}, and Sebastian Bader\IEEEauthorrefmark{1},~\IEEEmembership{Senior Member,~IEEE}
}

 \IEEEauthorblockA{\IEEEauthorrefmark{1}
 {Department of Computer and Electrical Engineering, Mid Sweden University, Sundsvall, Sweden}\\
\IEEEauthorblockA{\IEEEauthorrefmark{2}Instituto de Investigación en Señales, Sistemas e Inteligencia Computacional, sinc(i), FICH-UNL/CONICET, 3000 Santa Fe, Argentina}
}

}

\markboth{Preprint submitted to IEEE Open Journal of Instrumentation \& Measurement}
{}

\IEEEtitleabstractindextext{%
\begin{abstract}
Conveyor belts are crucial in mining operations by enabling the continuous and efficient movement of bulk materials over long distances, which directly impacts productivity.
While detecting anomalies in specific conveyor belt components has been widely studied, identifying the root causes of these failures, such as changing production conditions and operator errors, remains critical.
Continuous monitoring of mining conveyor belt work cycles is still at an early stage and requires robust solutions.
Recently, an anomaly detection method for duty cycle operations of a mining conveyor belt has been proposed.
Based on its limited performance and unevaluated long-term proper operation, this study proposes two novel methods for classifying normal and abnormal duty cycles.
The proposed approaches are pattern recognition systems that make use of threshold-based duty-cycle detection mechanisms, manually extracted features, pattern-matching, and supervised tiny machine learning models. The explored low-computational models include decision tree, random forest, extra trees, extreme gradient boosting, Gaussian naive Bayes, and multi-layer perceptron.
A comprehensive evaluation of the former and proposed approaches is carried out on two datasets.
Both proposed methods outperform the former method \textcolor{black}{in anomaly detection}, with the best-performing approach being dataset-dependent.
The heuristic rule-based approach achieves the highest \textcolor{black}{F1-score} in the same dataset used for algorithm training, with 97.3\% for normal cycles and 80.2\% for abnormal cycles. 
The ML-based approach performs better on a dataset including the effects of machine aging, \textcolor{black}{with an F1-score} scoring 91.3\% for normal cycles and 67.9\% for abnormal cycles.
Implemented on two low-power microcontrollers, the methods demonstrate efficient, real-time operation with energy consumption of 13.3 and 20.6 \textmu J during inference.
These results offer valuable insights for detecting mechanical failure sources, supporting targeted preventive maintenance, and optimizing production cycles.
\end{abstract}

}

\maketitle
\copyrightnotice

\IEEEdisplaynontitleabstractindextext

%
\IEEEpeerreviewmaketitle

\section{Introduction}
\label{s1}
\IEEEPARstart{T}{he} 
advancement in Industry~4.0 has transformed industrial operations, redefining manufacturing and production processes by harnessing the power of automation and digitalization. 
This evolution takes the name of Mining~4.0 in the specialized mining sector.
Mining~4.0 strategically integrates cutting-edge technologies such as the Internet of Things, machine learning~(ML), big data analytics, and cyber-physical systems~\cite{RAHMAN2023100822} to improve exploration, extraction, processing, and transportation~\cite{en16031427}.
Mining companies also invest in these tools as part of their commitment to safety, sustainability, and resource efficiency.
An aim is to reduce their environmental footprint by minimizing energy and water waste and adopting eco-friendly practices. By doing this, these companies aim to decrease operational risks, ensure compliance with environmental regulations, and align with global sustainability goals~\cite{BAI2020107776}.

Mining conveyor belts are vital for transporting materials over long distances and rough terrain. 
These belts allow for continuous and efficient movement of bulk materials, playing a critical role in optimizing productivity and minimizing downtime~\cite{ANDREJIOVA2016400}. 
The reliable operation of conveyor belts is paramount, as unexpected failures can lead to significant production interruptions, safety risks, and substantial economic losses. 
Therefore, detecting faults and anomalies in conveyor systems is essential. It helps to reduce unexpected downtime, prevent further damage, and maintain efficient operational workflows~\cite{10.1007/978-3-319-97490-3_61}.

Conveyor belt systems commonly feature various sensors to monitor their operational status and detect early warning signs of mechanical issues. 
Researchers have developed numerous methodologies using classic ML and deep learning to detect specific faults, such as idler and pulley defects~\cite{9459901}, belt damages~\cite{9387320}, reduced belt thickness~\cite{KIRJANOWBLAZEJ2023112744}, and belt deviation~\cite{ZHANG2022132575}. 
These approaches leverage data from vibration, thermal, X-ray, acoustic, radio-frequency, and camera sensors to identify these critical anomalies~\cite{ZHANG2023112735}.
While considerable attention has been paid to detecting failures in individual components, finding the root causes of failures is of utmost importance.
Failures may occur due to changing production conditions, incorrect control settings, operator errors, or equipment malfunctions. This highlights the need for robust, continuous operation monitoring of conveyor belts~\cite{7314930}.

In various industrial applications, such as predictive maintenance, manufacturing, infrastructure, and energy management, time-series data are employed to detect anomalies in operational productive processes using techniques like statistical analysis, data trend evaluation, and process mining~\cite{SHI2024103165,10379639}.
However, many existing methods are computationally intensive or require complex and costly sensor installations, which limits their feasibility in dynamic real-world environments. Due to these constraints, detecting anomalies in the mining sector remains especially challenging.


For continuous conveyor belt monitoring, Chiong~\emph{et~al.}~\cite{doi:10.1142/9789813143180_0005} applied unsupervised clustering to detect anomalies in pump cycle times, while Cheng~\emph{et~al.}~\cite{app13053244} analyzed coal mine communication data to identify operational anomalies and cyber threats. 
Anomaly deviations in a short belt were also studied under controlled laboratory conditions using inertial measurement units and a small dataset~\cite{10561167}.
Recently, Martinez-Rau~\emph{et~al.}~\cite{10636584} developed a method for identifying normal and abnormal duty cycle operations in an industrial mining conveyor belt.
However, this method demonstrates challenges in recognizing abnormal cycles and lacks an in-depth evaluation of machine drift over long periods of time, associated with wear or maintenance.
Furthermore, while the method demonstrated real-time operation on low-power microcontrollers (MCUs), its energy consumption was only estimated theoretically.

This work proposes two alternative lightweight pattern-recognition approaches.
The methods leverage manually extracted feature sets and a TinyML classifier~\cite{10433185} to identify the internal operation modes. 
In the first approach, pattern-matching rules analyze the sequence of internal operation modes to classify duty cycles as normal or abnormal. 
The second approach replaces these heuristic rules with another TinyML classifier trained to detect anomalous cycles while accounting for potential misclassifications in internal operation modes.
Six supervised classifiers are evaluated: decision tree (DT), random forest (RF), extra trees (ET)~\cite{geurts2006extremely}, extreme gradient boosting (XGB)~\cite{10.1145/2939672.2939785}, Gaussian naive Bayes (NB), and multi-layer perceptron (MLP) neural network~\cite{hastie2009elements}.
The best-performing models are deployed on two low-power MCUs for real-time anomaly detection.

The main contributions are the following:
(\emph{i}) The study presents two novel low-computational pattern recognition approaches for identifying abnormal conveyor belt usage in real-time on resource-constrained devices. 
The performance is evaluated and compared with a former method.
(\emph{ii}) 
The approaches are tested for machine drift over time on an additional dataset acquired more than a year after training.
(\emph{iii}) The approaches are optimized using quantization and implemented on two low-power MCUs, measuring energy consumption and memory usage.

The structure of the remaining parts of the article is as follows: the operation of the mining conveyor belt is detailed in Section~\ref{s2}. Section~\ref{s3} describes the data collection, the proposed approaches, the conducted experiments, and the evaluation methodology. Section~\ref{s4} presents the results and their interpretation. Finally, the conclusion and future research lines are discussed in Section~\ref{s5}.

\section{Conveyor Belt Operation}
\label{s2}
A hydraulic conveyor belt system operates by harnessing fluid power to drive the movement of the belt through hydraulic components. 
It is a highly efficient continuous transport mechanism, capable of moving heavy loads over long distances.
\textcolor{black}{In this system, an electric motor, powered by a starter or frequency converter, activates a hydraulic pump.
The pump pressurizes fluid, which is first filtered to remove contaminants, ensuring system longevity, and then circulated through a cooler to maintain optimal operating temperatures.
The pressurized fluid is directed through control valves, which regulate the flow and direction of oil to drive the hydraulic motors attached to the drive pulleys or rollers of the conveyor belt. 
These motors convert hydraulic energy into rotational motion, propelling the belt while the hydraulic fluid returns to a storage tank for recirculation.
\\
Operators control and monitor the control system via a control panel unit, connected to the electric motors, pumps, and valves.
The unit is equipped with sensors that continuously track and log key parameters. Operators use this information to adjust speed, direction, or load distribution of the conveyor belt, ensuring precise and efficient operation.
}

The specific conveyor belt used in this study can work in four different operation modes:
\begin{itemize}
    \item \emph{Off}: the system is not powered.
    \item \emph{Idle}: the \textcolor{black}{hydraulic} pumps are running, but the belt remains stationary.
    \item \emph{Operational}: the \textcolor{black}{hydraulic} pumps and belt are active, but no load is transported.
    \item \emph{Active}: the \textcolor{black}{hydraulic} pumps and belt operate, and the belt is transporting material.
\end{itemize}

The transitions between these modes can be aggregated to form two distinct types of operational cycles:
\begin{itemize}
    \item \emph{Normal} operation refers to the typical and expected use of the conveyor system. This cycle begins with a transition from \emph{Idle} to \emph{Operational}, initiated by the operator to start belt movement. When material is loaded onto the belt, the system shifts to \emph{Active} mode. The system remains in this mode while the load is transported. Once the load is cleared, two scenarios can occur: (\emph{}i) the operator deactivates the belt, returning the system to \emph{Idle} mode, or (ii) the belt continues moving without a load, transitioning from \emph{Active} back to \emph{Operational} until it is turned off and returns to \emph{Idle}. Thus, the full sequence of a normal operation cycle is either \emph{Idle-Operational-Active-Operational-Idle} or \emph{Idle-Operational-Active-Idle}.
    \item \emph{Abnormal} operation involves deviations from normal usage patterns, leading to energy waste and excessive wear on mechanical components. These cycles begin with a transition from a static belt position (\emph{Off} or \emph{Idle}) to a moving position (\emph{Operational} or \emph{Active}), followed by a return to a static position. Any sequence that differs from a normal cycle is classified as abnormal. For instance, the sequence \emph{Idle-Operational-Idle} is considered abnormal because the belt moves but does not transport any load. Another example of an abnormal cycle is \emph{Idle-Operational-Active-Operational-Active-Operational-Idle"}, which indicates that the load was not transported continuously, but in two portions, leading to inefficient operation.
\end{itemize}

\section{Materials and Methods}
\label{s3}
\subsection{Datasets}
\textcolor{black}{
The conveyor belt system used in this study operates under a control scheme that tries to maintain a fixed belt speed irrespective of the material load. 
As the load on the conveyor varies, the torque demand on the hydraulic motor changes proportionally, resulting in corresponding fluctuations in hydraulic pressure. 
These pressure variations serve as a proxy for assessing real-time system workload and mechanical stress. 
The system is instrumented with three sensors: an incremental encoder, and two hydraulic pressure transducers. 
The encoder, mounted on the hydraulic motor shaft, measures the motor speed in revolutions per minute (rpm) at high resolution (i.e., thousands of pulses per revolution).
The high-pressure transducer is positioned at the pump outlet or hydraulic motor inlet and measures the pressure of the fluid that drives the motor in bar. 
This pressure directly reflects the load conditions, as higher loads require greater pressure to maintain constant belt speed. 
The low-pressure transducer, placed at the pump inlet or on the motor return line to the tank, monitors return-line pressure in bar to ensure unrestricted fluid flow back to the reservoir. 
Elevated low-pressure values may indicate issues such as filter clogging or cooler obstruction. 
Both pressure sensors communicate with the control system via 4–20~mA analog signals, while the encoder is connected directly. 
Sensor data are sampled at 10~Hz by the control system for setpoint updates and local logging. 
For long-term monitoring and storage efficiency, the system averages each sensor’s readings over one-minute intervals and transmits these values to a secure central server. 
This study utilized two datasets retrieved through a web-based interface, comprising these one-minute interval records and enabling analysis of the conveyor belt’s operational parameters over time.}

\begin{figure}[tb!]
      \centering
      \includegraphics[width=1.0\columnwidth]{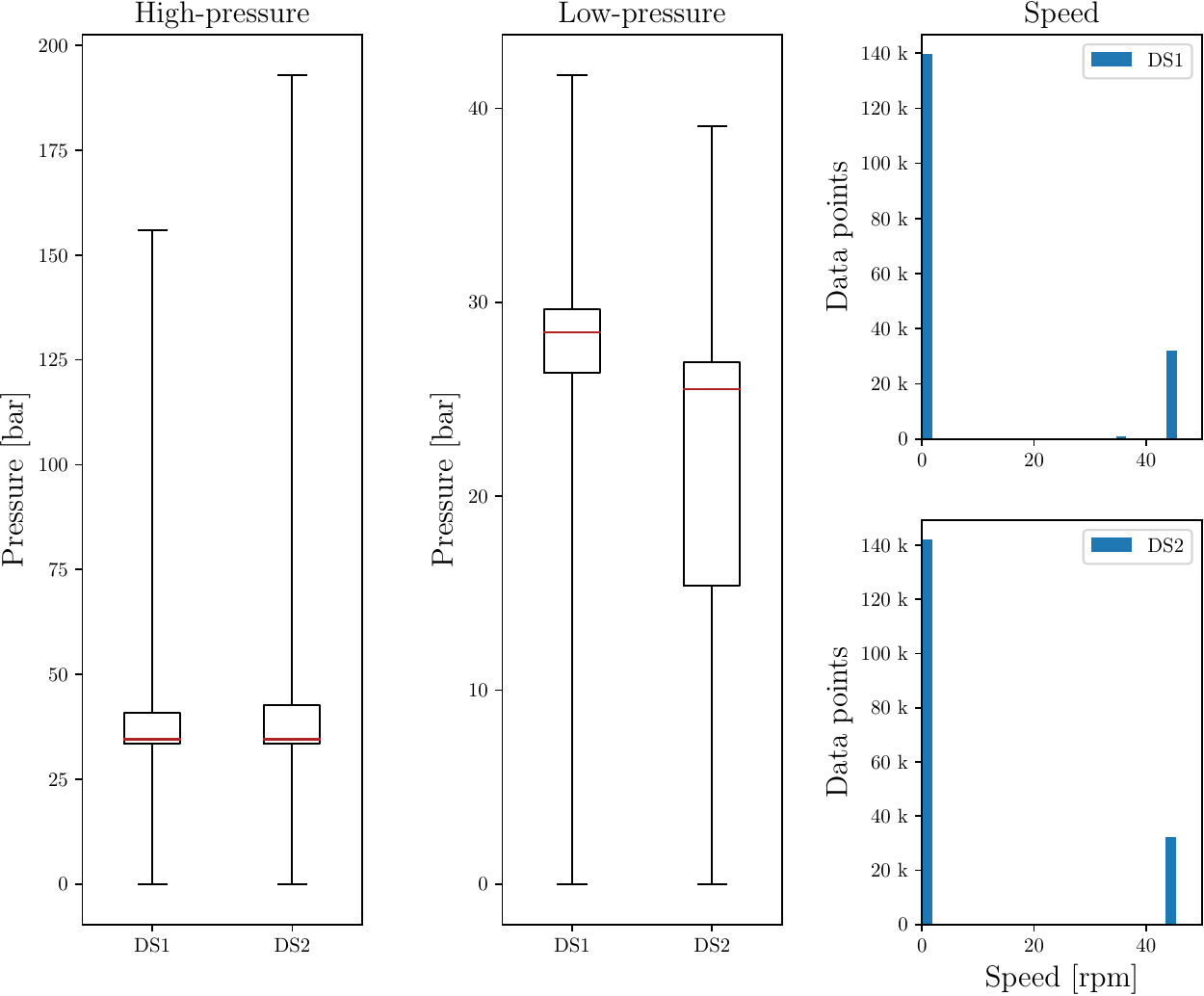}
      \caption{Box plots and histograms of the sensed input variables in DS1 and DS2.}
      \label{fig_boxplot}
\end{figure}

\textcolor{black}{
The first dataset (DS1) was collected over four months (June 2021, October 2021, January 2022, and April 2022), ensuring coverage of all four seasons and a broad temperature range (-26.1 to 26.0~°C).
The second dataset (DS2) was acquired from the same conveyor belt during June, August, October, and December 2023, with temperatures ranging from -29.0 to 26.0~°C.
These temperature variations significantly influenced oil viscosity, thereby affecting system pressure levels.
Each dataset comprises approximately 170,000 samples, where each sample represents one-minute averaged measurements of speed, high-pressure, and low-pressure values.}
Figure~\ref{fig_boxplot} shows the distribution of the three sensed variables in DS1 and DS2: high-pressure, low-pressure, and speed.
Whisker values in the box plots represent the minimum and maximum values.
Both datasets exhibit similar spreads for most high-pressure values, although DS2 spans a wider range.
DS1's low-pressure values span a greater range than those of DS2, with notable quartile differences.
This difference could be caused by machine wear and maintenance.
Speed values cluster around 0 and 45~rpm in both datasets when the conveyor belt is stopped or moving, respectively. 
Variations between these values occur during belt acceleration or deceleration.

\textcolor{black}{
Hydraulic system experts analyzed the waveform patterns of the data using the DEEPCRAFT™ Studio software (Infinion Technologies AG, Neubiberg, Germany) to determine the start and end timestamps, and the ground truth labels for the operation modes and duty cycles, described in Section~\ref{s2}.
Table~\ref{table_sample_dutycycle} presents the number of normal and abnormal duty cycles in each dataset.
Due to the extensive time consumption of the labeling task, operation mode labels are available exclusively on DS1. 
These labels were 1-minute segmented to match the three sensed variables, and their distribution is presented in Table~\ref{table_samples}.}
DS1 was used to develop and evaluate the proposed approaches, while DS2 was used to assess the generalization and robustness of the approaches in a different dataset.

\begin{table}[t]
\caption{Duty cycle samples distribution in DS1 and DS2.}
\label{table_sample_dutycycle}
\centering
\begin{tabular}{ccc}
\toprule
Dataset & Samples (percentage) & Class \\ \midrule
\multirow{2}{*}{DS1} & 496 (82.3\%) & Normal \\
 & 107 (17.7\%) & Abnormal \\
 \midrule
\multirow{2}{*}{DS2} & 751 (79.5\%) & Normal \\
 & 194 (20.5\%) & Abnormal \\ \bottomrule
\end{tabular}
\end{table}

\begin{table}[t]
\caption{Operation modes samples in DS1.}
\label{table_samples}
\centering
\begin{tabular}{cc}
\toprule
Samples (percentage) & Class  \\ \midrule
10,566 (6.2\%)       & \emph{Off}         \\
123,225 (72.7\%)     & \emph{Idle}        \\
11,454 (6.8\%)       & \emph{Operational} \\
24,299 (14.3\%)      & \emph{Active}      \\ \bottomrule
\end{tabular}
\end{table}

\subsection{Algorithms}
\label{s3b}
This study compares a state-of-the-art and two novel approaches to identify anomalies in conveyor belt operations. The approaches classify the duty cycle usages of a conveyor belt applying pattern recognition techniques.

\begin{figure*}[t]
\centering
    \begin{subfigure}{0.6\textwidth}
        \centering
        \includegraphics[width=\columnwidth]{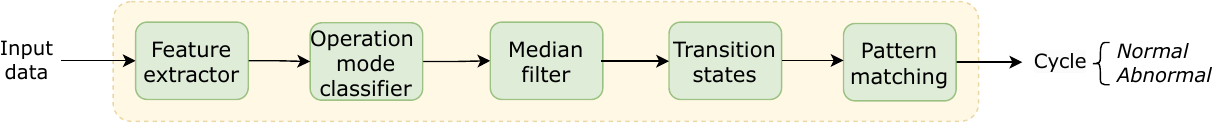}
        \caption{}
        \label{fig_approach1}
    \end{subfigure}
    \vfill
    \begin{subfigure}{0.6\textwidth}
        \centering
        \includegraphics[width=\columnwidth]{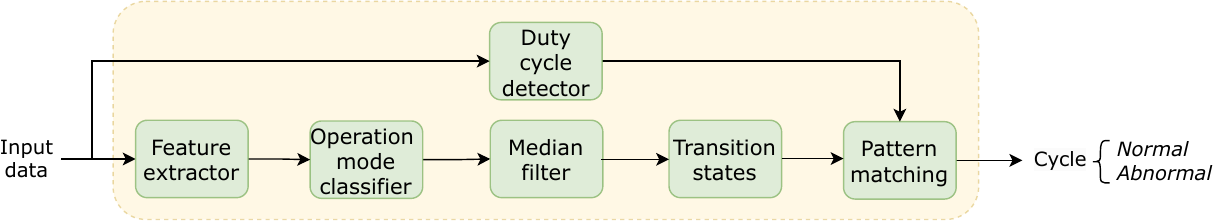}
        \caption{}
        \label{fig_approach2}
    \end{subfigure}
    \vfill
    \begin{subfigure}{0.6\textwidth}
        \centering
        \includegraphics[width=\columnwidth]{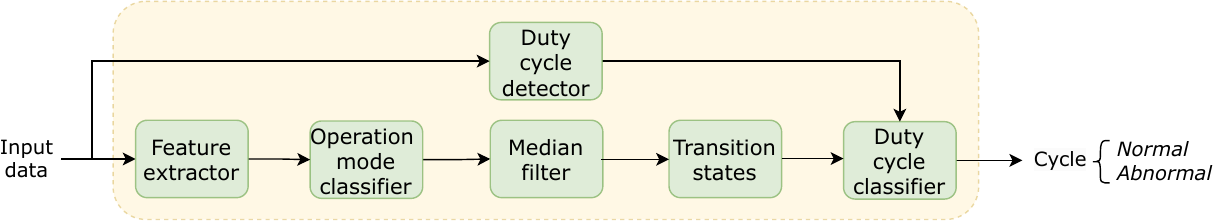}
        \caption{}
        \label{fig_approach3}
    \end{subfigure}
    \caption{Flow diagram of the proposed approaches: (a) \emph{Approach-1}, (b) \emph{Approach-2}, and (c) \emph{Approach-3}.}
    \label{fig_approaches}
\end{figure*}

\begin{enumerate}
    \item \emph{Approach-1} proposed by Martinez-Rau \textit{et~al.}~\cite{10636584}, consists of four consecutive stages (Fig.~\ref{fig_approach1}) and serves as a baseline. The algorithm takes as input the three 1-min sensed variables (motor shaft rotational speed, and the high- and low-pressure of the hydraulic motor), which are used to extract a set of 12~features. Four features consist of the three input values plus the differential pressure between the high- and low-pressure. 
    \textcolor{black}{Differential pressure determines the motor torque output.}
    The remaining eight features are derived from the former four features. These are third- and fifth-order moving averages, which effectively capture fast and slow variations in the input variables.
    The complete feature set feeds an ML model to classify the operation modes labeled every 1 min. Since variations in the operation modes typically occur at lower frequencies, a third-order median filter smooths the predicted labels to reduce label fragmentation. Finally, transitions in the filtered operation mode labels are analyzed through pattern matching to detect and classify normal and abnormal duty cycles.
    \textcolor{black}{The main drawback of this method is that duty-cycle detection depends on correct operation mode classification and good temporal alignment of the pattern matching analyzer.}
    \textcolor{black}{\emph{Approach-2} offers an alternative solution by decoupling the detection and classification tasks. Instead of relying solely on operation mode classification, this method first analyzes input speed values to identify the start and end of a duty cycle.}
    Based on the speed values distribution (Fig.~\ref{fig_boxplot}), an empirical threshold of 5~rpm is chosen to determine the boundaries of the duty cycles.
    Once these boundaries are identified, the remaining steps in \emph{Approach-2} follow a similar process to those in \emph{Approach-1}.
    \item \emph{Approach-3} builds on the concept of \emph{Approach-2} but introduces an ML model to classify duty cycles instead of relying on the pattern-matching rules used in the previous approaches (Fig.~\ref{fig_approach3}). 
    By employing an ML-based classifier, \emph{Approach-3} offers the advantage of generating smoother decision boundaries, which can help mitigate the impact of potential misclassifications of the operation modes.
    In this approach, each transition mode label links to an input of the duty-cycle classifier. 
    However, since the number of transitions within a duty cycle can vary, a value of 20~inputs was defined arbitrarily. 
    This value exceeds the maximum observed transitions (15) in both datasets.
    Entries not associated with a transition mode label are assigned a different label.
\end{enumerate}

The operation mode and the duty-cycle classifiers are evaluated with different combinations of supervised ML algorithms: DT, RF, ET, XGB, NB, and MLP.

\subsection{Experimental Setup}
The experiments were implemented in Python 3.8.10, utilizing the sklearn 1.2.2 package for tree-based and NB algorithms, and TensorFlow 2.13.0 for the MLP. 

\begin{table*}[tb]
\centering
\caption{Main properties of the MCU platforms}
\label{table_mcu}
\begin{tabular}{@{}ccccccccc@{}}
\toprule
MCU &
  Processor &
  Test board &
  CPU frequency &
  SRAM &
  Flash &
  FPU &
  Current consumption &
  Voltage \\ \midrule
nRF52840~\cite{ltc3600} &
  ARM Cortex M4 &
  \begin{tabular}[c]{@{}c@{}}Arduino Nano BLE \\ Sense 33 Lite\end{tabular} &
  64 MHz &
  256 kB &
  1 MB &
  \checkmark &
  52 \textmu A/MHz &
  3.3 V \\
RP2040~\cite{rp2040} &
  Dual-core ARM Cortex M0+ &
  Raspberry Pi Pico &
  133 MHz &
  256 kB &
  2 MB &
  $\times$ &
  75 \textmu A/MHz &
  3.3 V \\ \bottomrule
\end{tabular}
\end{table*}

The three approaches were trained and evaluated in the first experiment using DS1. 
The operation mode classifiers were trained and tested using a leave-one-month-out cross-validation strategy. In each iteration, three months of data were used for training, while the remaining month served for testing.
Synthetic oversampling and random undersampling techniques were applied using the SMOTE algorithm to balance the training set~\cite{he2008adasyn}.
In each leave-one-month-out iteration, 100\% of the minority \emph{Off} and \emph{Operational} classes were added in the training set, while 80\% of the majority \emph{Idle} classes were removed.
The gridsearchCV method of sklearn was used for hyperparameter tuning to optimize model performance.
This method applied 5-fold cross-validation on the balanced training set to train and evaluate the model with fixed hyperparameters. 
The macro-averaged F1-score per class was computed in each fold. Then, a final fold-averaged F1-score was computed to select the best hyperparameter values.
The following hyperparameters were evaluated during the optimization.
The maximum depth of the DT was set with a restriction of 10, 20, 30, 40, and 50 splits, as well as without restriction. In all cases, the Gini impurity and the Shannon entropy were considered as criteria for the quality split~\cite{hastie2009elements}. For the tree-based ensemble algorithms, the number of trees was set to 10, 25, and 50, with maximum tree depths of 4, 6, 8, and~10. The MLP architecture employed a single hidden layer, with the number of neurons varying between 4 and~15. 
The number of neurons in the input layer was fixed at~12 (number of features) and the number of neurons in the output layer was~4 (number of operation mode labels).
Training was conducted using stochastic gradient descent, with learning rates of 0.1, 0.01, and~0.001.
The other hyperparameters kept the default values defined in the respective Python packages.

In \emph{Approach-1} and \emph{Approach-2}, the predicted operation mode labels were used to detect and classify duty cycles based on heuristic rules (Fig.~\ref{fig_approaches}). Similar to the operation mode classifier, the duty-cycle classifier in \emph{Approach-3} was trained 
and tested using leave-one-month-out cross-validation.
In each iteration, the number of abnormal duty cycles was augmented using the SMOTEN algorithm to have the same number of normal and abnormal cycle samples in the training set~\cite{he2008adasyn}.
Hyperparameter optimization was performed in the same way as when training the operation mode classifier.
The operation mode classifier was trained first, then the duty-cycle classifier was trained and evaluated using the predicted operation mode labels across the four months. 
This approach allows the duty-cycle classifier to learn and adapt to potential misclassifications made by the operation mode classifier.

The generalization capabilities of the proposed approaches were evaluated in the second experiment, where the operation mode and duty-cycle classifiers were trained on the entire DS1 dataset and tested on DS2. 
During training, the same hyperparameter optimization and class balancing techniques were applied as in the first experiment.

A key factor to consider was the real-time operational requirements for anomaly monitoring in conveyor belts. To assess resource-constrained hardware requirements, the best-performing models in each dataset were implemented in C code. The best-performing ML models from the previous two experiments were converted to C code using the emlearn 0.19.3 package\footnote{\href{https://github.com/emlearn/emlearn/tree/0.19.3}{https://github.com/emlearn/emlearn/tree/0.19.3}} for tree-based and NB models, and LiteRT\footnote{\href{https://ai.google.dev/edge/litert}{https://ai.google.dev/edge/litert}} for the MLP. 
Similar to the second experiment, the models were trained on DS1 and tested on DS2.
The energy consumption, memory usage, and algorithm performance were analyzed using ML models with 32-bit float resolutions and post-training 8-bit integer full model quantization. The compiled C~code, using arm-none-eabi-g++, was deployed on two MCU boards: the Raspberry Pi Pico and the Arduino Nano Sense 33~BLE Lite. Table~\ref{table_mcu} provides the specifications of these boards.

The temporal alignment between the reference and recognized duty cycles has to be considered to evaluate duty cycle classification performance. The assessment was carried out using the sed\_eval package~\cite{Mesaros2021-vh}, which applies three criteria for correct duty cycle recognition: (\textit{i}) the class of the classified cycle must match the reference cycle, (\textit{ii}) the start timestamp must lie within the reference onset~$\pm$~a predefined tolerance, and (\textit{iii}) the end timestamp must lie within the reference offset~$\pm$~a predefined tolerance. A tolerance value of 202.75 seconds was chosen, representing 25\% of the minimum duration of a normal duty cycle. A smaller tolerance enforces stricter alignment, while a larger tolerance allows for greater overlap between the reference and recognized labels.

\begin{figure*}[tbh!]
      \centering
      \includegraphics[width=0.70\textwidth]{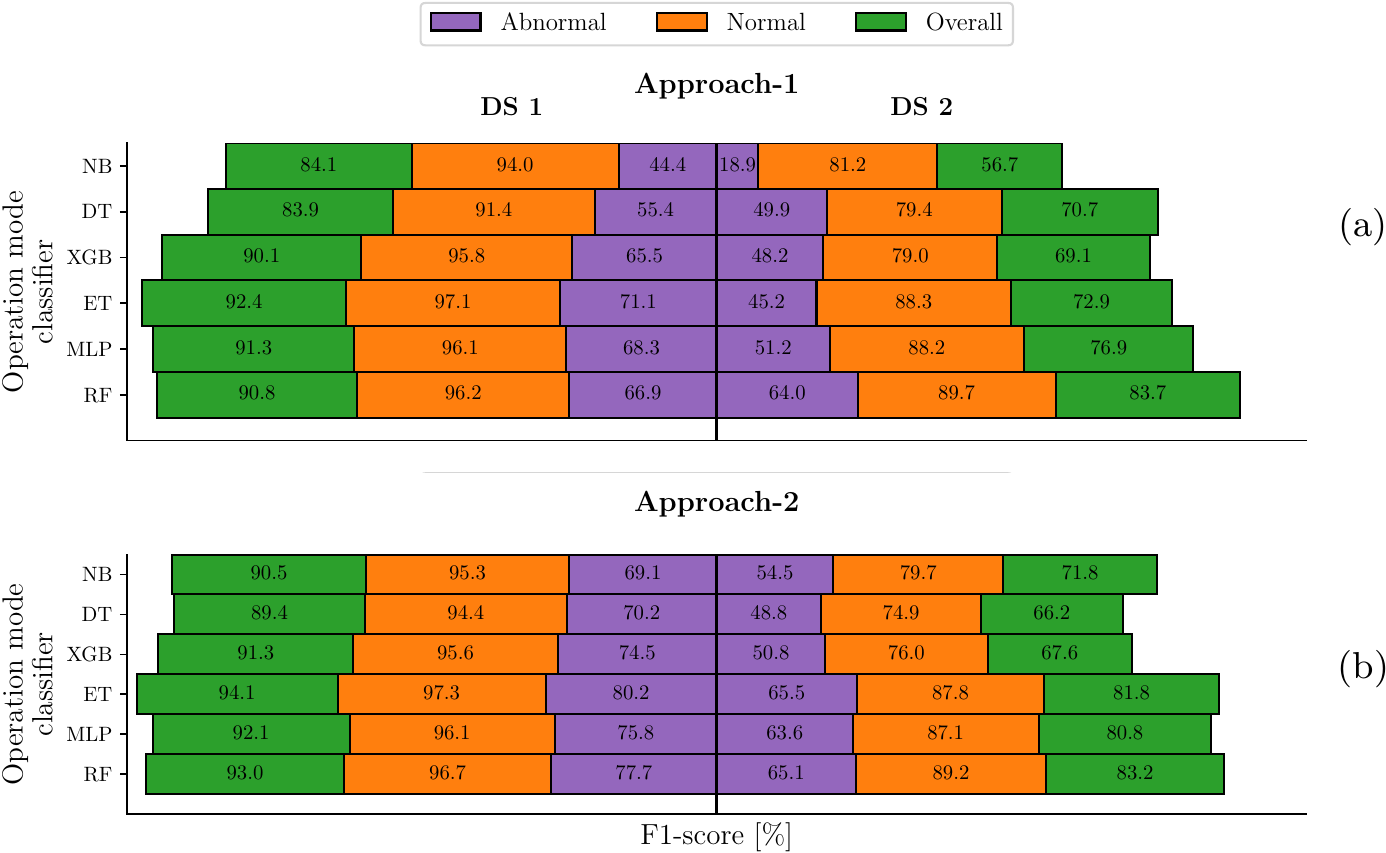}
      \caption{Diverging bar chart of the accumulative performance of (a) \emph{Approach-1} and (b) \emph{Approach-2}, using different operation mode classifiers in DS1~vs.~DS2.}
      \label{fig_exp}
\end{figure*}

Performance was measured using the F1-score for each class~$i$ of duty cycles (normal and abnormal), calculated as:

\[
F1-score_{i}=\frac{2 * TP_{i}}{2 * TP_{i} + FP_{i}+FN_{i}}
\]

\noindent where $TP_{i}$ counts the number of correctly classified instances of class $i$, $FP_{i}$ counts the number of recognized instances of class $i$ that do not exist in the ground truth (insertions), and $FN_{i}$ counts the number of ground truth instances of class $i$ that are not recognized (deletions). An overall micro-average F1-score was also computed to aggregate the total false positives, false negatives, and true positives across both classes~\cite{sokolova2009systematic}.
Except for the deployment on MCUs, each train/test experiment was repeated with different random seeds ten times to ensure robustness. For \emph{Approach-3}, 100~iterations were performed by combining the ten repetitions of the duty-cycle classifiers with the ten repetitions of the operation mode classifiers. The final metrics reported are the averages over all repetitions.

\section{Results and Discussion}
\label{s4}

\begin{figure}[t!]
      \centering
      \includegraphics[width=1.0\columnwidth]{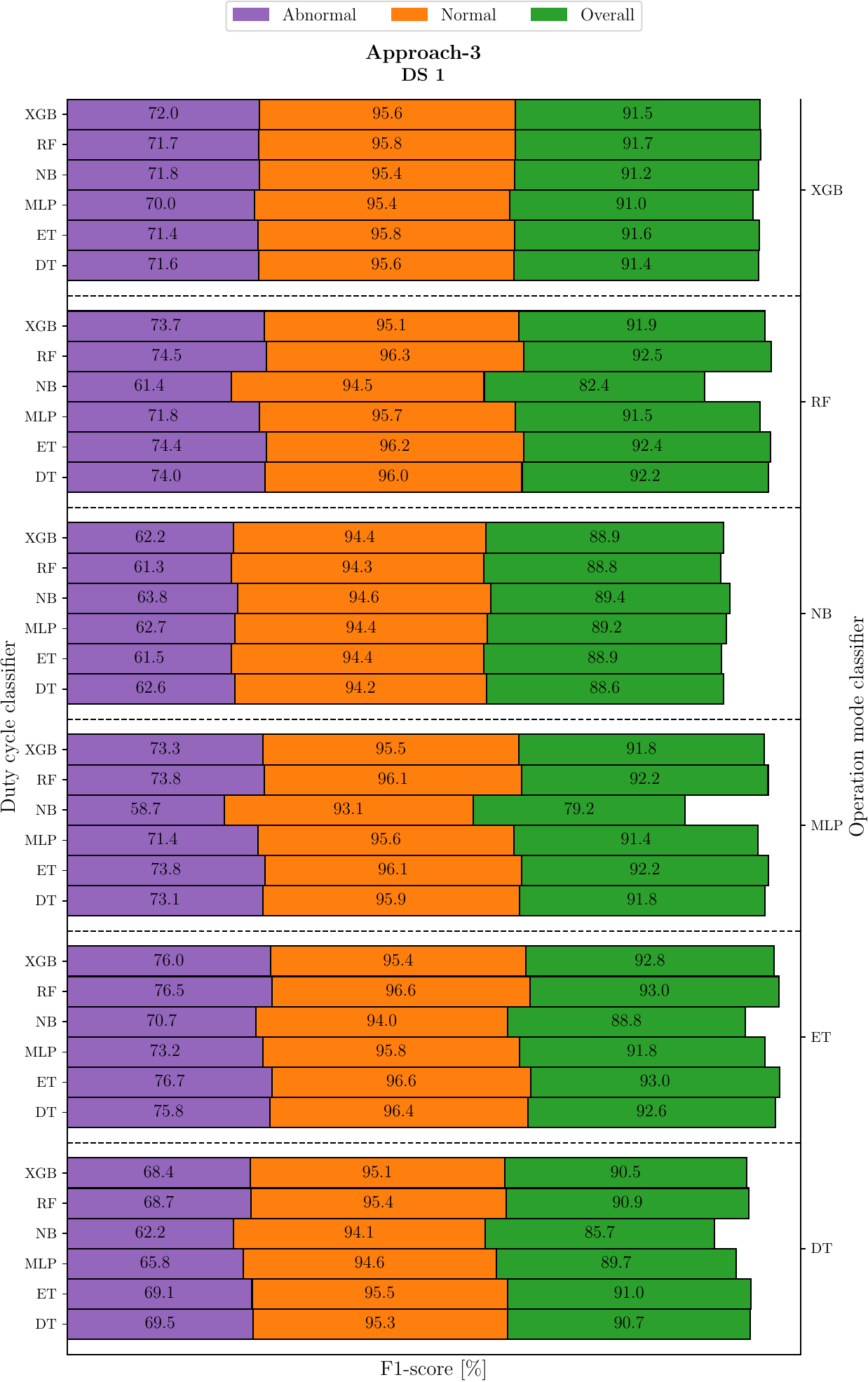}
      \caption{Grouped bar chart of the accumulative performance of \emph{Approach-3} combining different operation mode and duty-cycle classifiers in DS1.}
      \label{fig_exp3_ds1}
\end{figure}

\begin{figure}[t!]
      \centering
      \includegraphics[width=1.0\columnwidth]{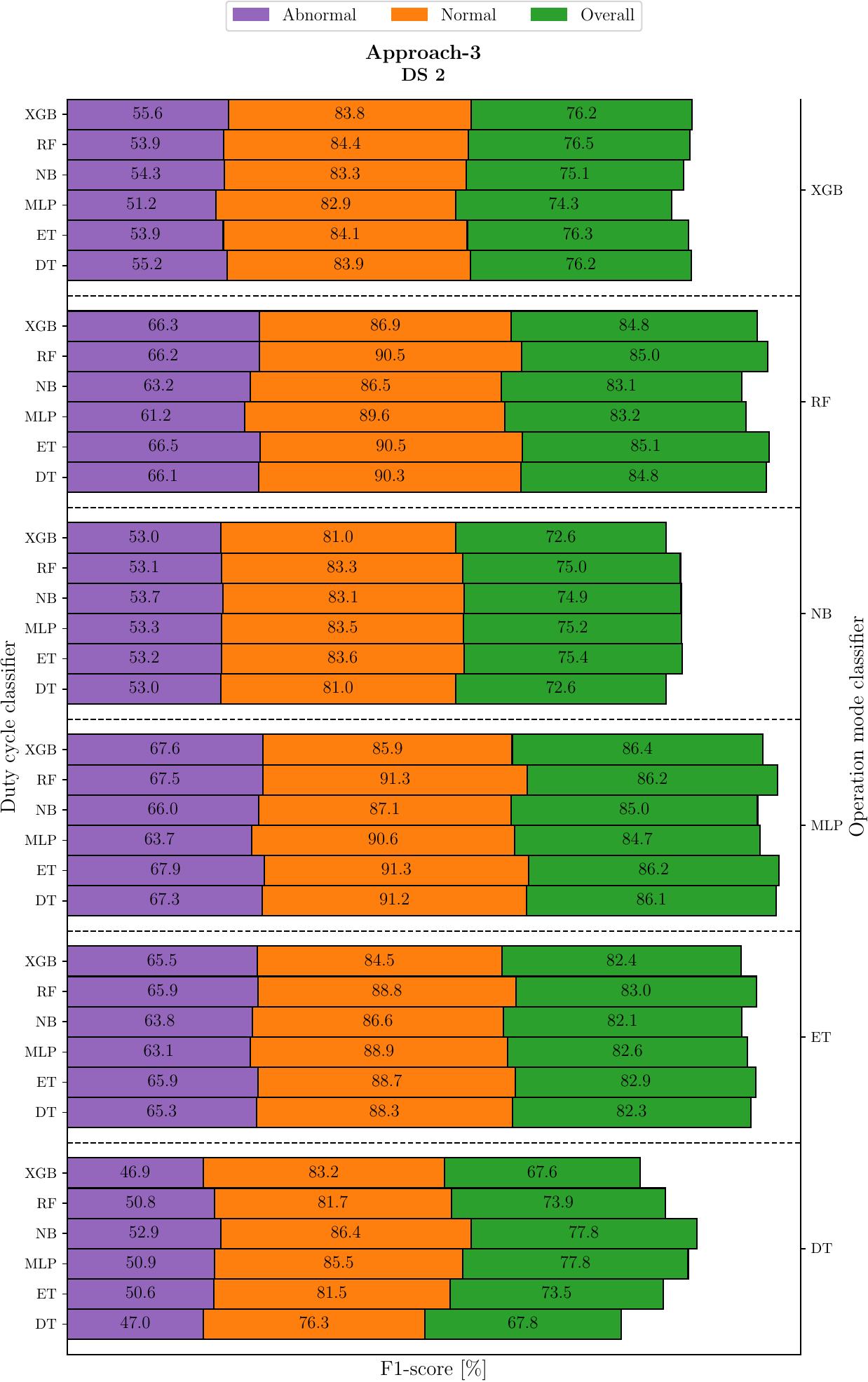}
      \caption{Grouped bar chart of the accumulative performance of \emph{Approach-3} combining different operation mode and duty-cycle classifiers in DS2.}
      \label{fig_exp3_ds2}
\end{figure}

\begin{table}[t!]
\centering
\caption{Average F1-score for the duty cycle detection in each dataset.}
\label{table_detection}
\begin{threeparttable}[t]
\begin{tabular}{lccc}
\toprule
Approach & \textcolor{black}{Detection method} & DS1 & DS2 \\ \midrule
\multirow{6}{*}{\emph{Approach-1}\tnote{*}} & NB & 89.3 & 61.6 \\
 & DT & 94.7 & 97.9 \\
 & XGB & 94.4 & 96.4 \\
 & ET & 95.0 & 82.4 \\
 & MLP & 92.5 & 87.8 \\
 & RF & 94.2 & 98.0 \\
 \midrule
\emph{Approach-2} & \multirow{2}{*}{\textcolor{black}{TBD\tnote{**}}} & \multirow{2}{*}{\textbf{97.6}} & \multirow{2}{*}{\textbf{99.2}} \\
\emph{Approach-3} &  &  &  \\ \bottomrule
\end{tabular}
\begin{tablenotes} \footnotesize
    \item[*]\textcolor{black}{Duty-cycle detection depends on the operation mode classifier.}
     \item[**]\textcolor{black}{Threshold-based detection (TBD) independent of the classifier.}
   \end{tablenotes}
\end{threeparttable}
\end{table}

\subsection{Recognition Performance}
The detection and classification performance of the approaches using different ML models is presented in terms of F1-scores, averaged over repeated experiments with varying seeds of initialization.
Table~\ref{table_detection} shows the detection rates for operation cycles across approaches. In \emph{Approach-1}, correct cycle detection relies on accurate classification of operation modes, yielding consistent performance above 89.3\% across all ML models in DS1. In DS2, the detection performance varies across ML models. While RF, DT, and XGB models show improvement over DS1, MLP, ET, and NB experience declines. In contrast, \emph{Approach-2} and \emph{Approach-3} detect cycles independently of the operation mode classifier, using threshold-based detection \textcolor{black}{(TBD)} (see Fig.~\ref{fig_approach2} and Fig.~\ref{fig_approach3}). This results in 2.5\% to 8.2\% improvements in DS1 and 1.3\% to 37.6\% in DS2 over \emph{Approach-1}.

Figures~\ref{fig_exp} to~\ref{fig_exp3_ds2} illustrate the duty cycle classification results for the three approaches in both datasets, revealing significant performance differences. 
DS1 consistently yields better results than DS2, reflecting the limited generalization capabilities of the approaches. This performance gap may stem from variations in the data range between the two datasets. These variations could be caused by differences in the transported materials, which may influence signal characteristics and separability. These findings suggest distinct complexities and class separability challenges between the two datasets.
Across all approaches, abnormal cycle classification is consistently more challenging and yields lower F1-scores than normal cycles. This is due to (\emph{i}) class imbalance and (\emph{ii}) the restrictive two possible operation mode sequence combinations in normal cycles, compared to the diverse sequences in the abnormal class (see section~\ref{s3b}).

In DS1, \emph{Approach-1} with the ET classifier achieves the highest performance, with an overall F1-score of 92.4\%, distinguishing well between normal (97.1\%) and abnormal (71.1\%) cycles (Fig.~\ref{fig_exp}a). The performance ranking is ET $>$ MLP $>$ RF $>$ XGB $>$ NB $>$ DT, with overall F1-scores decreasing from 92.4\% to 83.9\%. In DS2, however, RF outperforms other models with an overall F1-score of 83.7\%, attaining 64.0\% for abnormal and 89.7\% for normal cycles, suggesting superior generalization. Other models show a decline in abnormal cycle classification performance by 7.2-25.9\% and 8.0-16.4\% for normal cycles.

\textcolor{black}{\emph{Approach-2} (Fig.~\ref{fig_exp}b) and \emph{Approach-3} (Fig.~\ref{fig_exp3_ds1} and Fig.~\ref{fig_exp3_ds2})} demonstrate substantial improvements over \emph{Approach-1} across both datasets. 
In DS1, \emph{Approach-2} maintains a comparable F1-score for normal cycles and shows a 7.5-24.6\% improvement in abnormal cycle classification. ET again leads with an overall F1-score of 94.0\%, achieving 97.3\% for normal and 80.2\% for abnormal cycles, improving overall performance by 1.7\% and abnormal classification by 9.0\% over \emph{Approach-1}. In DS2, normal cycle classification with \emph{Approach-2} models is 0.5-4.1\% lower than \emph{Approach-1}. However, abnormal classification improves by 12.4\%, 20.2\%, and 35.6\% when using the MLP, ET, or NB models, respectively. This boosts the overall F1-scores by 3.9\%, 8.9\%, and 15.0\%. \emph{Approach-2} with the ET achieves the highest abnormal cycle F1-score, while RF excels in normal cycle and overall F1-scores.
\textcolor{black}{These results confirm that the pattern matching technique for classifying duty-cycles is effective, but is not suitable for their detection.}

For \emph{Approach-3}, which combines separate classifiers for operation modes and duty cycles, the highest performance in DS1 is using ET for both classifiers, reaching an overall F1-score of 93.0\% (96.6\% for normal, 76.7\% for abnormal) (Fig.~\ref{fig_exp3_ds1}). However, these results do not surpass the best performance obtained in \emph{Approach-2} using ET. Interestingly, in DS2, the combination of MLP for operation mode and ET for duty cycle classification yields the best performance in \emph{Approach-3}, with an overall F1-score of 86.2\% (91.3\% for normal, 67.9\% for abnormal) (Fig.~\ref{fig_exp3_ds2}). This result, the highest in DS2 across all approaches, suggests the dual-classifier structure may better address the complexity of DS2.
These findings emphasize the importance of aligning classification methods with dataset characteristics. They also highlight the potential benefits of sophisticated architectures in complex classification scenarios.
The proposed methods are specific to a hydraulic conveyor belt.
Their adaptability to new machines will depend on whether they allow monitoring of the same variables.
TinyML models will need to be retrained on new machines with new configurations (physical dimensions or conveying speed).

\begin{table*}[th!]
\centering
\caption{Comparison between the selected approaches deployed on the MCUs.}
\label{table_deployments}
\begin{threeparttable}[t]
\begin{tabular}{ccccccccccc}
\toprule
\multirow{4}{*}{Approach} & \multirow{4}{*}{\begin{tabular}[c]{@{}c@{}}Quantization\\ resolution\end{tabular}} & \multicolumn{3}{c}{F1-score {[}\%{]}} & \multicolumn{3}{c}{Arduino Nano BLE Sense 33~Lite} & \multicolumn{3}{c}{Raspberry Pi Pico} \\ \cmidrule(lr){3-5} \cmidrule(lr){6-8} \cmidrule(lr){9-11}
 &  & Abnormal & Normal & Overall & \begin{tabular}[c]{@{}c@{}}Energy\\ consumption\\ {[}\textmu J/inference{]}\end{tabular} & \begin{tabular}[c]{@{}c@{}}Flash \\ memory \\ {[}kB{]}\end{tabular} & \begin{tabular}[c]{@{}c@{}}RAM\\ {[}kB{]}\end{tabular} & \begin{tabular}[c]{@{}c@{}}Energy\\ consumption\\ {[}uJ/inference{]}\end{tabular} & \begin{tabular}[c]{@{}c@{}}Flash \\ memory \\ {[}kB{]}\end{tabular} & \begin{tabular}[c]{@{}c@{}}RAM\\ {[}kB{]}\end{tabular} \\ \midrule
\multirow{2}{*}{\emph{Approach-2}\tnote{1}} & 32-bit float & 64.4 & 87.8 & 81.4 & 18.8 & 265.9 & 43.1 & 82.7 & 209.5 & 10.4 \\
 & 8-bit integer & 64.9 & 87.9 & 81.6 & 13.7 & 142.4 & 43.0 & 13.3 & 116.4 & 10.3 \\
\multirow{2}{*}{\emph{Approach-3}\tnote{2}} & 32-bit float & 65.5 & 89.2 & 83.5 & 26.5 & 254.9 & 47.7 & 40.9 & 201.3 & 15.2 \\
 & 8-bit integer & 66.4 & 89.7 & 84.1 & 24.4 & 242.0 & 47.4 & 20.6 & 191.6 & 15.0 \\ \bottomrule
\end{tabular}
\begin{tablenotes} \footnotesize
     \item[1] Using ET \textcolor{black}{with 50 trees} as operation mode classifier.
     \item[2] \textcolor{black}{Using MLP with 12 neurons in the hidden layer as operation mode classifier and ET with 50 trees as duty cycle classifier.}
   \end{tablenotes}
\end{threeparttable}
\end{table*}

\subsection{MCU Energy Requirements}

The best-performing classifier-dependent approaches for each dataset are selected for deployment on MCUs using 32-bit float and 8-bit integer representation. 
These include \emph{Approach-2}, which uses an ET classifier for operation mode classification, and \emph{Approach-3}, which combines an MLP for operation mode classification and ET for duty cycle classification. 

Quantizing to 8-bit does not impact duty cycle detection accuracy, which remained at 99.2\% as observed in earlier experiments (Table~\ref{table_detection}).
Table~\ref{table_deployments} illustrates the balance between classification performance in DS2 and hardware resource usage. 
\emph{Approach-3} achieves superior classification rates for normal, abnormal, and overall classes compared to \emph{Approach-2}, consistent with previous results. 
Interestingly, both approaches show improved classification performance when quantized to 8-bit, likely due to reduced quantization error.

Energy consumption calculations focused on data representing duty cycles, ensuring that all system stages are active (Fig.~\ref{fig_approaches}), capturing the peak energy consumption.
Reported energy consumption reflects the end-to-end computation for each 1-min input data, averaged across DS2. 
After completing the inference, the idle time can be utilized for additional tasks. These include MCU peripheral management, conveyor control, or putting the MCU into sleep mode.

\textcolor{black}{
The Arduino Nano BLE Sense~33 Lite, which features a floating-point computation unit (FPU), consumes less energy than the Raspberry Pi Pico when running the 32-bit float versions of the approaches. 
The effect of quantization to 8-bit integer on the Arduino Nano BLE Sense~33 Lite reduces energy consumption by 27.1\% for \emph{Approach-2} and 7.9\% for \emph{Approach-3}. 
This reduction occurs because operations performed by the FPU in the non-quantized versions are executed by the CPU in the quantized versions. 
The impact of quantization is more pronounced on the Raspberry Pi Pico, decreasing energy usage by 83.9\% for \emph{Approach-2} and 49.6\% for \emph{Approach-3}.
\\
The cycle classifier contributes minimally to overall energy consumption since it performs only a single inference at the end of each cycle. Thus, the key difference in energy usage between \emph{Approach-2} and \emph{Approach-3} arises from their distinct operational mode classifiers.
Notably, \emph{Approach-3} consumes more energy than \emph{Approach-2} on the Arduino Nano BLE Sense~33 Lite. This is because the ET classifier (\emph{Approach-2}) primarily relies on logical comparisons, which can be executed in a single clock cycle on the FPU, whereas the MLP classifier (\emph{Approach-3}) requires more complex operations related to activation functions.
Conversely, the opposite behavior is observed for the floating-point versions of the approaches on the Raspberry Pi Pico. For this MCU, the C code compiler inserts additional code to emulate floating-point operations using integer arithmetic, resulting in more overhead for ET than for the MLP.}

Quantizing \emph{Approach-2} reduces program memory usage by 46.4\% on the Arduino Nano BLE Sense 33 Lite and 44.4\% on the Raspberry Pi Pico, while \emph{Approach-3} presents reductions of 5.1\% and 4.8\%, respectively. 
Across both MCUs, 32-bit float and 8-bit integer versions required similar amounts of RAM, as both store data in 32-bit registers.
\textcolor{black}{
The low resource usage of the deployed methods in the MCUs suggests their compatibility with other general-purpose MCUs. Building on this efficiency,} we hypothesize that the power consumption of quantized approaches could be further reduced by using ultra-low-power 8-bit MCUs, such as the STM8L series MCUs~\cite{stm8}.

\section{Conclusion}
\label{s5}
This work contributes to Mining 4.0 by proposing two novel approaches for real-time anomaly detection in the continuous operation of hydraulic conveyor belt systems using MCUs. 
These pattern recognition methods combine manual feature extraction, threshold-based cycle detection, and supervised TinyML classifiers to accurately identify normal and abnormal duty cycles. 
By evaluating the performance of different approaches and ML classifiers across two distinct datasets, we identified the most effective methods for achieving accurate and generalizable classification. These methods are essential for conveyor belt system monitoring and predictive maintenance. 
The first \textcolor{black}{alternative (\emph{Approach-2})} performed better when trained and tested on the same dataset, achieving an F1-score of 97.3\% for normal and 80.2\% for abnormal cycles. In contrast, the second \textcolor{black}{alternative (\emph{Approach-3})} showed superior performance on an independent dataset, with F1-scores of 91.3\% for normal and 67.9\% for abnormal cycles.
This comparison highlights both the strengths and adaptability challenges of the methods across varying operational contexts.
Furthermore, both approaches outperformed the state-of-the-art method on both datasets, validating their suitability for real-world scenarios.

When deployed in two resource-constrained MCUs, the methods demonstrated stable performance with 32-bit float representation and even slight improvements with 8-bit integer quantization, resulting from minimal quantization error. 
Energy consumption remained low, at 13.3~\textmu J and 20.6~\textmu J per inference on a Raspberry Pi Pico, enabling seamless integration into conveyor control systems and potential for automated preventive actions.

Taking advantage of the fact that the deployed approaches in MCUs use a small amount of the available resources, future work will explore advanced feature engineering techniques to enhance discriminative power and robustness, thereby improving generalization across diverse operating conditions and datasets.
Additionally, improving duty cycle classification will be investigated by employing more complex models, such as recurrent neural networks or transformer-based models, to capture temporal dependencies better.
Another research direction involves addressing the limited availability of abnormal cycle data through few-shot learning, as well as adding new unlabeled data into the model development using unsupervised ML techniques, which may enhance anomaly detection without extensive labeled datasets.

\section*{Acknowledgment}
This research was financially supported by the Knowledge Foundation under grant NIIT 20180170.

\ifCLASSOPTIONcaptionsoff
  \newpage
\fi


\bibliographystyle{./IEEEtran}
\bibliography{bibliography}

\begin{thebibliography}{10}
\providecommand{\url}[1]{#1}
\csname url@samestyle\endcsname
\providecommand{\newblock}{\relax}
\providecommand{\bibinfo}[2]{#2}
\providecommand{\BIBentrySTDinterwordspacing}{\spaceskip=0pt\relax}
\providecommand{\BIBentryALTinterwordstretchfactor}{4}
\providecommand{\BIBentryALTinterwordspacing}{\spaceskip=\fontdimen2\font plus
\BIBentryALTinterwordstretchfactor\fontdimen3\font minus \fontdimen4\font\relax}
\providecommand{\BIBforeignlanguage}[2]{{%
\expandafter\ifx\csname l@#1\endcsname\relax
\typeout{** WARNING: IEEEtran.bst: No hyphenation pattern has been}%
\typeout{** loaded for the language `#1'. Using the pattern for}%
\typeout{** the default language instead.}%
\else
\language=\csname l@#1\endcsname
\fi
#2}}
\providecommand{\BIBdecl}{\relax}
\BIBdecl

\bibitem{RAHMAN2023100822}
M.~S. Rahman, T.~Ghosh, N.~F. Aurna, M.~S. Kaiser, M.~Anannya, and A.~S. Hosen, ``Machine learning and internet of things in industry 4.0: A review,'' \emph{Meas.: Sens.}, vol.~28, p. 100822, 2023.

\bibitem{en16031427}
O.~Zhironkina and S.~Zhironkin, ``Technological and intellectual transition to mining 4.0: A review,'' \emph{Energies}, vol.~16, no.~3, 2023.

\bibitem{BAI2020107776}
C.~Bai, P.~Dallasega, G.~Orzes, and J.~Sarkis, ``Industry 4.0 technologies assessment: A sustainability perspective,'' \emph{Int. J. Prod. Econ.}, vol. 229, p. 107776, 2020.

\bibitem{ANDREJIOVA2016400}
M.~Andrejiova, A.~Grincova, and D.~Marasova, ``Measurement and simulation of impact wear damage to industrial conveyor belts,'' \emph{Wear}, vol. 368-369, pp. 400--407, 2016.

\bibitem{10.1007/978-3-319-97490-3_61}
L.~Jurdziak, R.~Blazej, and M.~Bajda, ``Conveyor belt 4.0,'' in \emph{Proc. Intell. Syst. Prod. Eng. Maint.}, A.~Burduk, E.~Chlebus, T.~Nowakowski, and A.~Tubis, Eds.\hskip 1em plus 0.5em minus 0.4em\relax Cham: Springer International Publishing, 2019, pp. 645--654.

\bibitem{9459901}
R.~L. de~Moura, D.~Bibancos, L.~P. Barreto, A.~C. Fracaroli, and E.~Martinelli, ``Study case in mining industry: Monitoring rollers using embedded lorawan,'' in \emph{Proc. IEEE Int. Instrum. Meas. Technol. Conf. (I2MTC)}, 2021, pp. 1--5.

\bibitem{9387320}
O.~Salim, S.~Dey, H.~Masoumi, and N.~C. Karmakar, ``Crack monitoring system for soft rock mining conveyor belt using uhf rfid sensors,'' \emph{IEEE Trans. Instrum. Meas.}, vol.~70, pp. 1--12, 2021.

\bibitem{KIRJANOWBLAZEJ2023112744}
A.~Kirjanów-Błażej, L.~Jurdziak, R.~Błażej, and A.~Rzeszowska, ``Calibration procedure for ultrasonic sensors for precise thickness measurement,'' \emph{Meas.}, vol. 214, p. 112744, 2023.

\bibitem{ZHANG2022132575}
M.~Zhang, K.~Jiang, Y.~Cao, M.~Li, N.~Hao, and Y.~Zhang, ``A deep learning-based method for deviation status detection in intelligent conveyor belt system,'' \emph{J. Clean. Prod.}, vol. 363, p. 132575, 2022.

\bibitem{ZHANG2023112735}
M.~Zhang, K.~Jiang, Y.~Cao, M.~Li, Q.~Wang, D.~Li, and Y.~Zhang, ``A new paradigm for intelligent status detection of belt conveyors based on deep learning,'' \emph{Meas.}, vol. 213, p. 112735, 2023.

\bibitem{7314930}
Y.~Hu, Y.~Yan, L.~Wang, X.~Qian, and X.~Wang, ``Simultaneous measurement of belt speed and vibration through electrostatic sensing and data fusion,'' \emph{IEEE Trans. Instrum. Meas.}, vol.~65, no.~5, pp. 1130--1138, 2016.

\bibitem{SHI2024103165}
Y.~Shi, N.~Zhang, X.~Song, H.~Li, and Q.~Zhu, ``Novel approach for industrial process anomaly detection based on process mining,'' \emph{J. Process Control}, vol. 136, p. 103165, 2024.

\bibitem{10379639}
P.~Yan, A.~Abdulkadir, P.-P. Luley, M.~Rosenthal, G.~A. Schatte, B.~F. Grewe, and T.~Stadelmann, ``A comprehensive survey of deep transfer learning for anomaly detection in industrial time series: Methods, applications, and directions,'' \emph{IEEE Access}, vol.~12, pp. 3768--3789, 2024.

\bibitem{doi:10.1142/9789813143180_0005}
R.~Chiong, Z.~Hu, Z.~Fan, Y.~Lin, S.~Chalup, and A.~Desmet, ``A bio-inspired clustering model for anomaly detection in the mining industry,'' in \emph{Bio-Inspired Computing Models and Algorithms}.\hskip 1em plus 0.5em minus 0.4em\relax World Scientific, 2019, ch.~5, pp. 133--155.

\bibitem{app13053244}
Z.~Cheng, B.~Cui, and J.~Fu, ``Rethinking the operation pattern for anomaly detection in industrial cyber–physical systems,'' \emph{Appl. Sci.}, vol.~13, no.~5, 2023.

\bibitem{10561167}
S.~N. Matos, O.~F. Coletti, R.~Zimmer, F.~U. Filho, R.~C.~L. de~Carvalho, V.~R. da~Silva, J.~L. Franco, T.~V.~B. Pinto, L.~G.~D. de~Barros, C.~M. Ranieri, B.~E. Lopes, D.~F. Silva, J.~Ueyama, and G.~Pessin, ``Machine learning techniques for improving multiclass anomaly detection on conveyor belts,'' in \emph{Proc. IEEE Int. Instrum. Meas. Technol. Conf. (I2MTC)}, 2024, pp. 1--6.

\bibitem{10636584}
L.~S. Martinez-Rau, Y.~Zhang, B.~Oelmann, and S.~Bader, ``Tinyml anomaly detection for industrial machines with periodic duty cycles,'' in \emph{Proc. IEEE Sensors Appl. Symp. (SAS)}, July 2024, pp. 1--6.

\bibitem{10433185}
L.~Capogrosso, F.~Cunico, D.~S. Cheng, F.~Fummi, and M.~Cristani, ``A machine learning-oriented survey on tiny machine learning,'' \emph{IEEE Access}, vol.~12, pp. 23\,406--23\,426, 2024.

\bibitem{geurts2006extremely}
P.~Geurts, D.~Ernst, and L.~Wehenkel, ``Extremely randomized trees,'' \emph{Mach. Learn.}, vol.~63, pp. 3--42, 2006.

\bibitem{10.1145/2939672.2939785}
T.~Chen and C.~Guestrin, ``Xgboost: A scalable tree boosting system,'' in \emph{Proc. 22nd ACM SIGKDD Int. Conf. Knowl. Discov. Data Min.}\hskip 1em plus 0.5em minus 0.4em\relax New York, NY, USA: Association for Computing Machinery, 2016, p. 785–794.

\bibitem{hastie2009elements}
T.~Hastie, R.~Tibshirani, J.~H. Friedman, and J.~H. Friedman, \emph{The elements of statistical learning: data mining, inference, and prediction}.\hskip 1em plus 0.5em minus 0.4em\relax Springer, 2009, vol.~2.

\bibitem{ltc3600}
\emph{{nRF52840} Datasheet. Product Specification v1.11}, Nordic semiconductor, 2024.

\bibitem{rp2040}
\emph{{RP2040} Datasheet. A microcontroller by Raspberry Pi}, Raspberry Pi Ltd, 2024.

\bibitem{he2008adasyn}
H.~He, Y.~Bai, E.~A. Garcia, and S.~Li, ``Adasyn: Adaptive synthetic sampling approach for imbalanced learning,'' in \emph{Proc. IEEE Int. Joint Conf. Neural Netw}.\hskip 1em plus 0.5em minus 0.4em\relax IEEE, 2008, pp. 1322--1328.

\bibitem{Mesaros2021-vh}
A.~Mesaros, T.~Heittola, T.~Virtanen, and M.~D. Plumbley, ``Sound event detection: A tutorial,'' \emph{IEEE Signal Process. Mag.}, vol.~38, no.~5, pp. 67--83, 2021.

\bibitem{sokolova2009systematic}
M.~Sokolova and G.~Lapalme, ``A systematic analysis of performance measures for classification tasks,'' \emph{Inf. Process. Manag.}, vol.~45, no.~4, pp. 427--437, 2009.

\bibitem{stm8}
\emph{{STM8} 8‑bit MCU family. Jump to new record heights! Simply Smarter}, STMicroelectronics, 2018.

\end{thebibliography}

\begin{IEEEbiography}[{\includegraphics[width=1in,height=1.25in,clip,keepaspectratio]{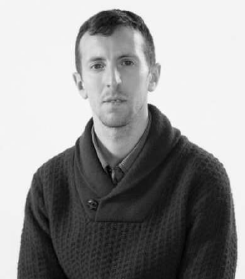}}]{Luciano Martinez-Rau} (Member, IEEE) received the Ph.D. degree in Engineering Science from National University of the Littoral, Argentina in 2022.
Since 2023, he is Postdoctoral Researcher at Mid Sweden University, Department of Computer and Electrical Engineering, Sweden. He is also an Adjunct Professor at the National University of the Littoral in Argentina.
His research interests include low-power embedded systems, machine learning, signal processing and on-device training.
\end{IEEEbiography}

\begin{IEEEbiography}[{\includegraphics[width=1in,height=1.25in,clip,keepaspectratio]{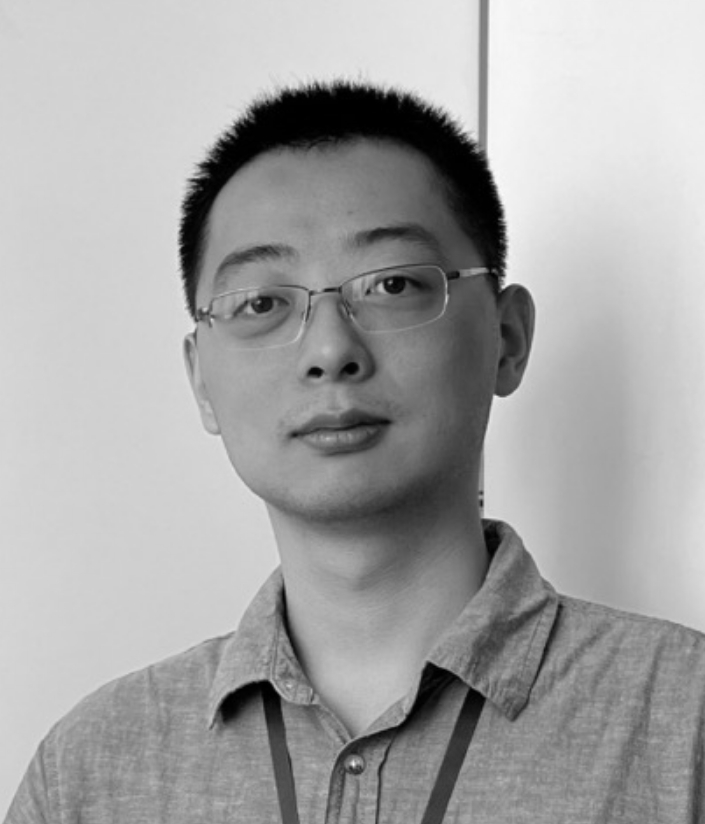}}]{Yuxuan Zhang}
(Graduate Student Member, IEEE) received the MSc degree in Embedded Systems Engineering from the University of Leeds, UK in 2019. Since 2021, he is a Ph.D. candidate of Electronics at the Department of Computer and Electrical Engineering, Mid Sweden University, Sweden. His research interests include machine learning \& signal processing on low-power resource-constrained embedded devices and on-device training.
\end{IEEEbiography}

\begin{IEEEbiography}[{\includegraphics[width=1in,height=1.25in,clip,keepaspectratio]{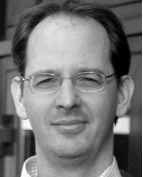}}]{Bengt Oelmann} received the Doctor of Technology degree in electronics from the Royal Institute of Technology, Stockholm, Sweden, in 2000. He is currently a Full Professor in electronics system design with Mid Sweden University, Sundsvall, Sweden. His current research interests include low-energy embedded system design, energy harvesting, and embedded sensor technology.
\end{IEEEbiography}

\begin{IEEEbiography}[{\includegraphics[width=1in,height=1.25in,clip,keepaspectratio]{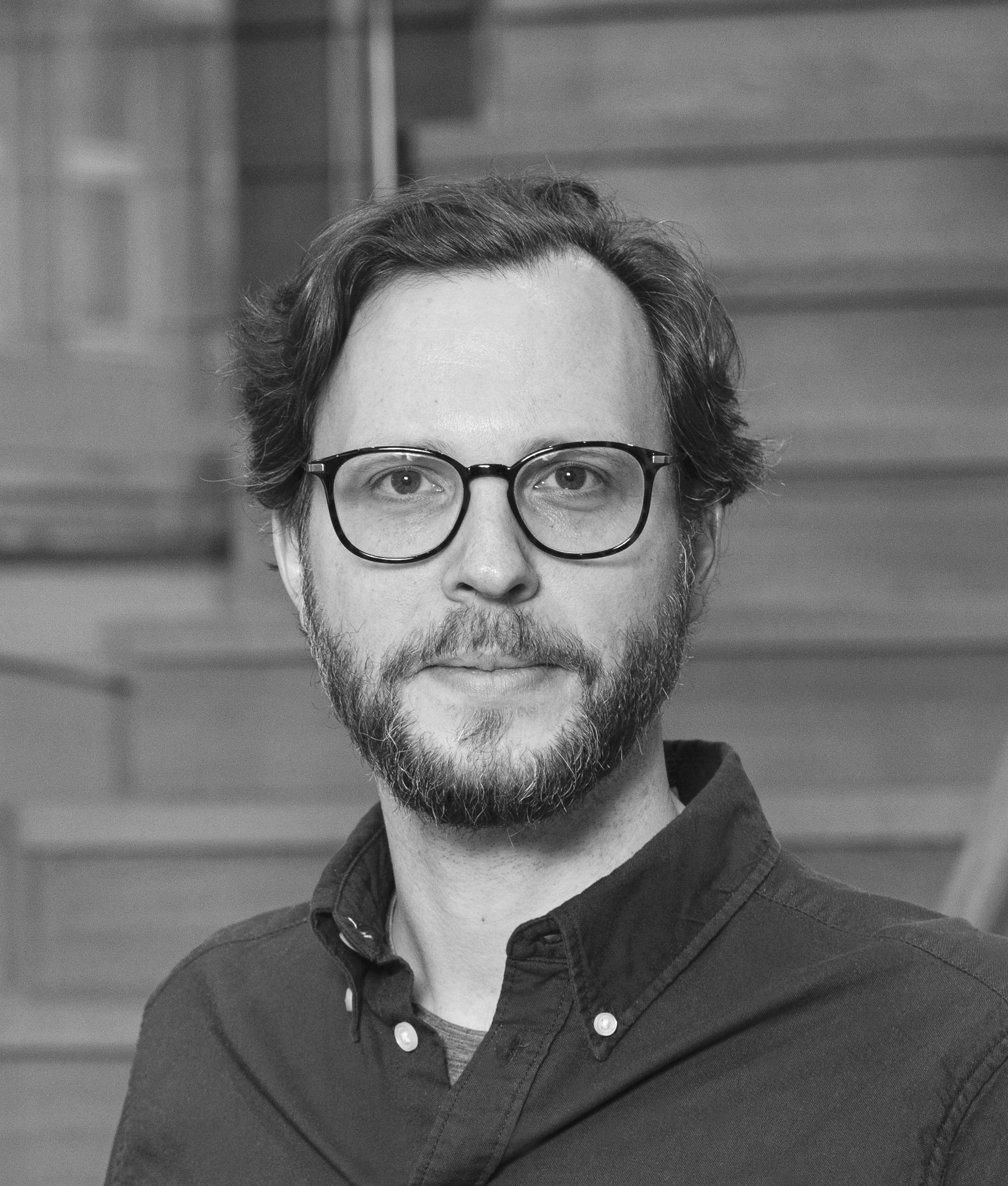}}]{Sebastian Bader} (Senior Member, IEEE) received the Ph.D. degree in electronics from Mid Sweden University, Sundsvall, Sweden, in 2013, and the Dipl.-Ing. degree from the University of Applied Sciences, Wilhelmshaven, Germany in 2008. He is currently an Associate Professor of embedded systems with the Department of Computer and Electrical Engineering, Mid Sweden University. His research interests focus on energy aspects of embedded systems, including energy harvesting, low-power sensing systems, and machine learning on resource-constrained devices.
\end{IEEEbiography}




\end{document}